\documentclass{bmvc2k}

\usepackage{multirow}
\usepackage{booktabs}
\usepackage{subcaption}
\usepackage{graphicx}
\usepackage{amsmath}
\usepackage{amssymb}
\usepackage{enumitem}

\title{MS-GAGA: Metric-Selective Guided Adversarial Generation Attack}

\newcommand\blfootnote[1]{%
  \begingroup
  \renewcommand\thefootnote{}\footnote{#1}%
  \addtocounter{footnote}{-1}%
  \endgroup
}

\addauthor{Dion J. X. Ho$^\dagger$}{dh3065@columbia.edu}{1}
\addauthor{Gabriel Lee Jun Rong$^\dagger$}{2301906@sit.singaporetech.edu.sg}{2}
\addauthor{Niharika Shrivastava$^\ast$}{a103955@singaporetech.edu.sg}{2}
\addauthor{Harshavardhan Abichandani$^\ast$}{harshavardhan@u.nus.edu}{3}
\addauthor{Pai Chet Ng}{paichet.ng@singaporetech.edu.sg}{2}
\addauthor{Xiaoxiao Miao}{xiaoxiao.miao@dukekunshan.edu.cn}{4}

\addinstitution{
 Columbia University\\
 New York, USA
}
\addinstitution{
 Singapore Institute of Technology\\
 Singapore
}
\addinstitution{
 Independent Researcher
}
\addinstitution{
 Duke Kunshan University\\
 China
}

\runninghead{Ho et al.}{Metric-Selective Guided Adversarial Generation Attack}

\begin{document}

\maketitle
\blfootnote{$^\dagger$,$^\ast$ Equal contribution.}

\begin{abstract}
We present \textbf{MS-GAGA} (Metric-Selective Guided Adversarial Generation Attack), a two-stage framework for crafting transferable and visually imperceptible adversarial examples against deepfake detectors in black-box settings.
In Stage 1, a dual-stream attack module generates adversarial candidates: MNTD-PGD applies enhanced gradient calculations optimized for small perturbation budgets, while SG-PGD focuses perturbations on visually salient regions. This complementary design expands the adversarial search space and improves transferability across unseen models.
In Stage 2, a metric-aware selection module evaluates candidates based on both their success against black-box models and their structural similarity (SSIM) to the original image. By jointly optimizing transferability and imperceptibility, MS-GAGA achieves up to 27\% higher misclassification rates on unseen detectors compared to state-of-the-art attacks.
\end{abstract}

\section{Introduction}
\label{sec:intro}

\subsection{Background and Motivation}

Rapid advances in artificial intelligence, particularly deep learning, have facilitated the creation of synthetic media known as "deepfakes". Modern deepfakes are often generated using sophisticated diffusion-based models \cite{ho2020denoising, rombach2022high} which iteratively refine synthetic media until it is virtually indistinguishable from authentic content. The quality of these generated images and videos has reached a level where they are highly realistic and difficult to distinguish from real face images, blurring the line between reality and fabrication \cite{K_2024_CVPR, chen2023text}. This has precipitated a host of severe societal and ethical challenges. Deepfakes have been weaponized to erode public trust, spread disinformation in political contexts, violate personal privacy through non-consensual content, and perpetrate widespread fraud \cite{10552098}.

In response to these growing threats, the development of accurate and reliable deepfake detection algorithms has become a critical and urgent challenge for researchers, policymakers, and industry stakeholders alike. Recent years have witnessed the emergence of a diverse range of detection approaches, broadly categorized into spatial, temporal, and adversarially robust methods. Spatial-domain methods, such as the XceptionNet-based detector \cite{rossler2019faceforensics} in the FaceForensics++ benchmark, leverage convolutional architectures to capture subtle pixel-level inconsistencies introduced during generation. Transformer-based models, including Vision Transformer (ViT) variants such as the Patch Transformer \cite{wang2022m2tr, guan2022delving} for Deepfake Detection, extend this capability by modeling global dependencies to identify manipulation artifacts. Temporal-domain approaches exploit frame-to-frame inconsistencies, with systems like Lip Forensics Detector (LFD) \cite{haliassos2021lips} analyzing lip-sync discrepancies, while multi-attentional video transformers detect temporal flicker and jitter patterns using cross-frame attention.

Despite the promising performance of deepfake detectors under normal conditions, a new and formidable challenge has emerged in the form of adversarial attacks \cite{szegedy2013intriguing}. In the context of deepfake detection, these attacks are designed to cause a detector, which would normally classify a synthetic image as "fake," to misclassify it as "real". This poses a significant threat to the reliability and effectiveness of deepfake detection systems, as it fundamentally undermines their ability to distinguish between genuine and manipulated content. The nature of these attacks represents a critical evolution in the threat landscape. Adversarial attacks do not rely on high-level, human-perceptible artifacts. Instead, they exploit the low-level vulnerabilities of the underlying neural network architecture itself \cite{hou2023evading, neekhara2021adversarial}. Recent work \cite{conf/bmvc/NgP24} has further shown that even pretrained face verification models, despite their strong identity representation capabilities, face challenges in reliably distinguishing genuine from manipulated identities. This signifies a far more sophisticated and difficult-to-counter threat.

This research is motivated by the critical need to understand the vulnerabilities of deepfake detection systems to adversarial attacks. Proactively developing and analyzing these attacks is a defensive strategy known as "red teaming" \cite{sinha2025can}. By intentionally exposing the weaknesses of existing detection models, researchers can devise more robust and resilient defences. The ultimate goal is to strengthen the deepfake detection algorithms' resistance to adversarial attacks and ensure the continued efficacy of media authentication systems. The study of adversarial attacks has a direct, practical application in developing countermeasures. Defense mechanisms such as adversarial training \cite{goodfellow2014explaining}, which involves exposing a model to adversarial examples during its training phase, have been shown to significantly enhance a model's robustness. This work aims to advance this field by creating a more realistic and powerful form of attack, thereby pushing the boundaries of what constitutes a robust defense. A significant portion of existing research on adversarial attacks against deepfake detectors has focused on the unrealistic "white-box" threat model \cite{liao2021imperceptible, jia2022exploring, hussain2021adversarial}, where the attacker has complete knowledge of the detector's architecture and parameters. While effective at demonstrating vulnerability, this approach does not represent the real-world scenario of a malicious actor facing a deployed "black-box" model. This creates a critical gap in the literature, which this research is designed to address. The focus on black-box attacks is thus a necessary step to produce solutions that are not just academically interesting but also practically relevant to securing AI systems in the real world.

\subsection{Problem Statement}

The primary objective is to devise and evaluate a novel framework for generating highly transferable, imperceptible adversarial attacks on deepfake image detection algorithms, specifically targeting the more realistic black-box threat model. This study aims to create attacks that can successfully evade a wide range of unknown, state-of-the-art detectors without requiring access to their internal architecture or parameters. Our methodology prioritizes maximizing the attack's transferability, thereby providing a robust method for stress-testing the security and reliability of deepfake detection systems in real-world scenarios.

We propose MS-GAGA \ref{ms-gaga}, a dual-stream adversarial framework for attacking black-box deepfake detectors. Our key contributions are two-fold:
\begin{enumerate}
    \item We introduce a novel dual-stream attack strategy that fuses diverse perturbation mechanisms, one driven by adaptive \textit{MNTD-PGD} \ref{mntd-pgd} optimization and the other guided by visual saliency \textit{SG-PGD} \ref{sg-pgd}, to expand the adversarial search space and improve robustness across models and content types.
    
     \item We develop a metric-aware selection module that jointly considers black-box misclassification success (as a proxy for transferability) and perceptual similarity (via SSIM \cite{wang2004image}), enabling dynamic, high-fidelity adversarial image selection tailored for varying input conditions.
\end{enumerate}


\section{Related Work}

\subsection{Deepfake Detection Systems}
Early advances in deepfake detection relied heavily on convolutional neural networks (CNNs), exemplified by the FaceForensics++ benchmark and its Xception-based classifier \cite{rossler2019faceforensics}. Such models exploit local texture artifacts and blending inconsistencies but tend to overfit to dataset-specific characteristics, leading to sharp performance drops in cross-dataset settings.

Frequency-domain methods, such as F3-Net \cite{qian2020thinking}, have sought to capture GAN-specific spectral patterns, offering improved robustness over RGB-only models. However, these approaches can be undermined by compression, post-processing, or novel generative architectures like diffusion models.

Region-specific detection approaches have been explored by \cite{calvert2024leveraging}. For instance, transfer learning has been leveraged to adapt detectors toward specific facial regions, improving sensitivity to localized manipulations. \cite{foo2025Emotions} empirically demonstrated that emotional expressions significantly impact facial recognition performance across diverse datasets, highlighting the importance of human-centric variability when defending against deepfake manipulations. Physiological prior–based detectors, such as DeepRhythm \cite{qi2020deeprhythm}, leverage subtle cues like remote photoplethysmography (rPPG) to detect manipulation. While effective in high-resolution, controlled and region-specific settings, they degrade under occlusion, variable lighting, or low resolution, and their applicability to still images remains limited.

Transformer-based and foundation model approaches (e.g., Swin Transformer hybrids \cite{liu2021swin} or CLIP-adapted detectors \cite{radford2021learning}) have demonstrated stronger generalization to in-the-wild forgeries by modeling global context and high-level semantics. Nonetheless, these architectures are data-hungry, computationally expensive, and still vulnerable to unseen generation pipelines.

Diffusion-aware methods, such as DiffusionFake \cite{chen2024diffusionfake}, reconstruct the input image using a frozen generative prior and detect inconsistencies between the original and reconstruction. While they excel at detecting manipulations from similar generative families, their reliance on prior-specific features limits performance against entirely novel architectures.

Finally, forensic approaches, including EXIF metadata consistency \cite{huh2018fighting} and sensor pattern noise analysis (Noiseprint/PRNU) \cite{cozzolino2019noiseprint}, provide interpretable detection cues but are brittle to common online transformations such as resizing, compression, and content re-hosting.

\subsection{Adversarial Attacks on Deepfake Detectors}

\paragraph{Taxonomy.}
Adversarial attacks are commonly classified by the attacker’s level of knowledge about the target model:
\begin{itemize}
    \item \textbf{White-Box Attacks:} The attacker has full access to the target model’s architecture, parameters, and gradients, enabling the creation of highly effective perturbations.
    \item \textbf{Black-Box Attacks:} The attacker has no internal knowledge of the model and can only query it for predictions, often exploiting transferability from a surrogate model.
    \item \textbf{Gray-Box Attacks:} The attacker has partial knowledge, such as the architecture but not the parameters.
\end{itemize}

Formally, the attacker seeks a perturbed input $x^{adv}$ such that a classifier $f$ misclassifies it, with $\delta = (x^{adv} - x)$ remaining imperceptible:
\begin{equation}
\underset{x^{adv}}{\arg\min} \ \mathcal{L}[f(x^{adv}), y] \quad \text{s.t.} \quad ||x^{adv} - x||_p \le \epsilon
\end{equation}
where $\epsilon$ controls perceptibility under an $L_p$ norm.

\paragraph{Recent Advances and Limitations.}
Despite improvements in accuracy, deepfake detection systems remain susceptible to adversarial perturbations. Carlini and Farid \cite{carlini2020evading} showed that imperceptible pixel-level noise can cause significant misclassification rates in state-of-the-art detectors, exploiting their over-reliance on high-frequency cues—making frequency-domain methods particularly vulnerable. Recent work such as AdvForensics \cite{neekhara2021adversarial} has proposed transferable perturbations across multiple detectors, raising concerns about model-agnostic vulnerabilities. Adaptive attacks that jointly optimize against multiple architectures \cite{LI2023103319} can bypass ensemble-based systems, once believed to be more robust. Beyond pixel-level perturbations, semantic adversarial attacks alter high-level image attributes such as lighting, pose, or background while maintaining visual plausibility \cite{qiu2019semanticadv}. These exploit distributional shifts rather than low-level noise vulnerabilities, impacting both CNN- and transformer-based detectors.

Nonetheless, adversarial methods face constraints. Many require white-box access, which is unrealistic in deployed systems. Transfer-based black-box attacks often degrade under compression or social media re-encoding, which remove or distort perturbations \cite{dziugaite2016studyeffectjpgcompression}. Moreover, adversarial training can improve robustness but frequently reduces clean-sample accuracy, making it a trade-off in practical detection systems \cite{zhang2019theoretically}. These concerns extend beyond academic benchmarks. Face recognition and generative AI pipelines are already being embedded in real-world applications such as personalized guest services in the hospitality sector \cite{herman2024integrating}, making adversarial robustness a pressing requirement.

\section{Proposed Methodology}
\label{ms-gaga}

\begin{figure}
\includegraphics[width=\textwidth]{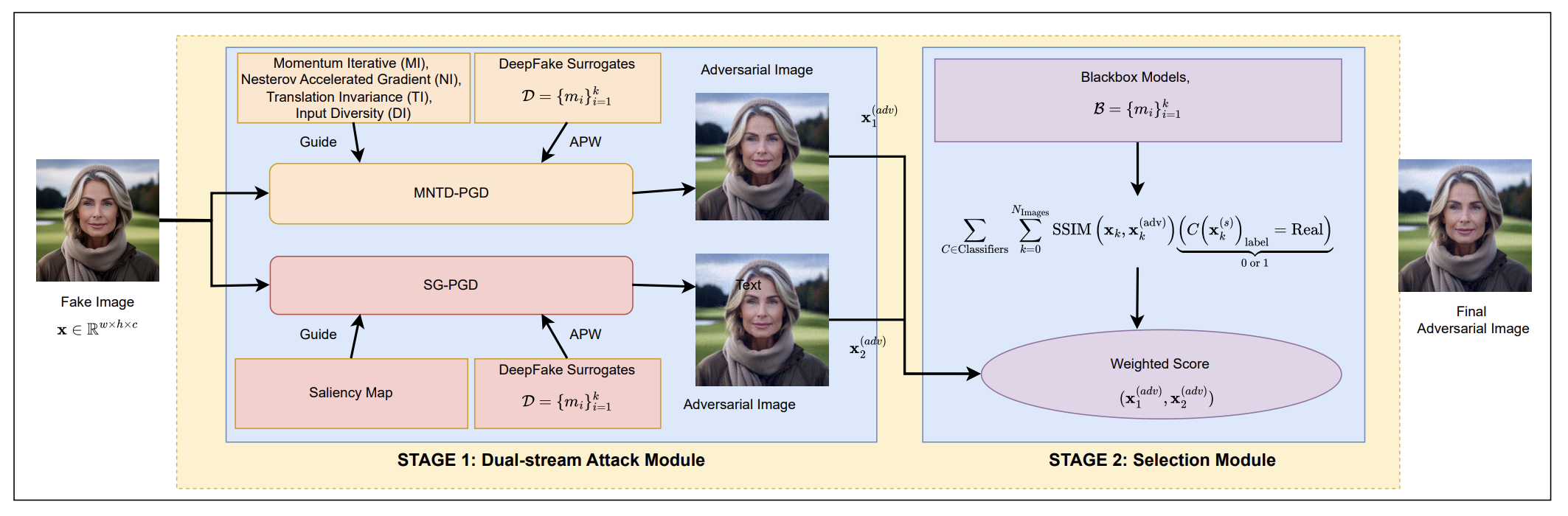}
\vspace{-0.2cm}
\caption{MS-GAGA framework with dual-stream attack module and a selection module.}
\label{fig:teaser}
\end{figure}

As shown in Figure \ref{fig:teaser}, MS-GAGA is a two-stage framework for generating transferable and imperceptible adversarial examples against black-box deepfake detectors. Stage 1 creates adversarial images with MNTD-PGD and SG-PGD, while Stage 2 selects the optimal image based on black-box misclassification and SSIM.

\subsection{Stage 1: Dual-stream Attack Module}
To enhance both adversarial transferability and perceptual imperceptibility, MS-GAGA introduces a dual-stream adversarial generation module. Given an input AI-generated fake image $\mathbf{x} \in \mathbf{R}^{H \times W \times C}$, the objective is to craft an adversarial counterpart $\mathbf{x}^{\text{adv}}$ that maintains high visual similarity to $\mathbf{x}$ while successfully misleading multiple black-box deepfake detectors. This stage generates two adversarial candidates by leveraging complementary strategies aimed at (1) maximizing transferability across unseen models and (2) preserving the structural integrity of the original image.

\subsubsection{Momentum-Nesterov-Translation-Diversity PGD (MNTD-PGD)}
\label{mntd-pgd}
The first stream extends the standard PGD framework with a suite of robust optimization strategies: (1) Momentum Iterative (M) \cite{Dong_2018_CVPR} accumulates past gradients to stabilize updates and escape local minima, (2) Nesterov Accelerated Gradient (N) \cite{lin2019nesterov} with a look-ahead step to refine gradient directionality, (3) Translation Invariance (T) \cite{Dong_2019_CVPR} applies Gaussian smoothing to improve robustness to spatial shifts, and (4) Input Diversity (D) \cite{xie2019improving} randomizes input size and padding at each iteration to prevent overfitting to specific surrogate models and boosting attack generalization across architectures.
Given an AI-generated fake image $\mathbf{x} \in \mathbf{R}^{H \times W \times C}$ and a deepfake detecting surrogate ensemble $\mathcal{D} = \{m_i\}_{i=1}^{k}$, the adversarial example $\mathbf{x}^{\text{adv}}$ is generated through iterative updates as follows:

\begin{equation}
\mathbf{x}^{\text{adv}}_{t+1} = \mathbf{x}^{\text{adv}}_t + \alpha \cdot \operatorname{sign}(g_t)
\end{equation}
\begin{equation}
g_t = \mu \cdot g_{t-1} + \frac{1}{k} \sum_{i=1}^k \nabla_{\mathbf{x}} \mathcal{L}_{\text{total}}(m_i(\mathbf{x}^{\text{adv}}_t), y)
\end{equation}
where $g_t$ is the momentum-accumulated gradient at iteration $t$, $\mu$ is the momentum decay factor, and $\alpha$ is the step size. 

We also incorporate an SSIM-regularized \cite{wang2004image} loss $\mathcal{L}_{\text{SSIM}}$ into our MNTD-PGD framework, directly optimizing for high visual similarity between original and adversarial images. Unlike traditional methods that only limit pixel changes, this approach explicitly shapes perturbations to be imperceptible. $\lambda_{\text{SSIM}}$ is a regularization term tuned to $0.3$. The total loss is defined as $\mathcal{L}_{\text{total}} = \mathcal{L}_{\text{misclassification}} + \mathcal{L}_{\text{SSIM}}$. 

\begin{equation}
\mathcal{L}_{\text{SSIM}} = \lambda_{\text{SSIM}} \cdot \left(1.0 - \operatorname{SSIM}(\mathbf{x}, \mathbf{x}^{\text{adv}})\right)
\end{equation}

Additionally, we employ several novel optimizations:
\begin{itemize}[noitemsep, topsep=0pt]
    \item \textbf{Adaptive Per-Surrogate Weighting (APW):} Instead of treating all surrogate models equally, APW dynamically assigns higher weights to those that are harder to fool or more confident in their predictions. This strategic allocation of gradient contribution focuses optimization on the most challenging surrogates, leading to more effective and transferable adversarial examples. By learning which models provide the most informative gradients, APW introduces a meta-optimization layer that enables more intelligent, resource-aware attack strategies.
    \item \textbf{Epsilon Search:} The standard, fixed-epsilon perturbations are replaced by an intelligent epsilon search to find the minimal perturbation needed for a successful attack. It starts with a coarse grid search, followed by binary search refinement between the last failed and first successful values. This process minimizes perturbations and inherently improves SSIM. In addition, the epsilon grid is scaled with the pixel variance of the image - a higher variance allows for larger epsilons.
    \item \textbf{Preprocessing:} For each image, we adjust contrast, brightness, and add Perlin noise \cite{perlin1985image} to create more "natural" looking final images. This was key to achieving a misclassification rate of $~99\%$ on our target models.
\end{itemize}

\subsubsection{Saliency-Guided PGD (SG-PGD)}
\label{sg-pgd}
To maximise the attack budget ($\epsilon$), we made use of saliency maps \cite{simonyan2013deep} to identify and target the most influential pixels in an image. Given an AI-generated fake image $\mathbf{x} \in \mathbf{R}^{H \times W \times C}$ and a surrogate model $m \in \mathcal{D}$, we first compute a saliency map by taking the gradient of the loss function $\mathcal{L}$ with respect to the input image, $M_{saliency} = \left| \nabla_{\mathbf{x}} \mathcal{L}(m(\mathbf{x}), y) \right|$. 
This mask isolates the most important regions, and the adversarial perturbation is subsequently multiplied by it. The update rule for creating the adversarial image, $x^{adv}$ at each step is modified to:

\begin{equation}
    \mathbf{x}^{\text{adv}}_{t+1} = \mathbf{x}^{\text{adv}}_{t} + \alpha \cdot \text{sign}\left( \nabla_{\mathbf{x}} \mathcal{L}(m(\mathbf{x}^{\text{adv}}_{t}), y) \right) \odot M_{\text{saliency}}
\end{equation}
Here, $\alpha$ is the step size and $\odot$ represents element-wise multiplication.

\subsection{Stage 2: Selection Module}
While Stage~1 generates two adversarial candidates ($x^{adv}_{MNTD\text{-}PGD}$ and $x^{adv}_{SG\text{-}PGD}$), not all perturbations are equally effective or perceptually similar. Simply choosing one arbitrarily risks sacrificing either transferability or imperceptibility. To address this, we design a metric-aware \emph{Selection Module} that evaluates each candidate according to both (i) its success in fooling black-box detectors and (ii) its structural similarity to the original image. 

Formally, given a set of classifiers $C \in \mathcal{C}$ and an input batch of $N_{images}$ samples, we compute a joint score that multiplies attack success with perceptual quality. For each adversarial image $x_k^{adv}$, the Structural Similarity Index (SSIM) between the original $x_k$ and $x_k^{adv}$ serves as a proxy for visual fidelity, while the classification outcome provides a binary measure of attack success. The cumulative score is defined as:
\begin{align}
score &= \sum_{C \in \text{Classifiers}}\sum_{k=1}^{N_{images}}SSIM\left(x_k, x_k^{adv}\right)\underbrace{\left(C\left(x_k^{adv}\right)_\text{label} = \text{Real}\right)}_{0 \text{ or } 1} 
\label{eqn: score}
\end{align}
where adversarial examples are rewarded only when they are both \emph{successful} (fool the classifier) and \emph{imperceptible} (high SSIM). Finally, the Selection Module compares the two streams and retains the candidate with the higher score:
\begin{align}
x^{final} &= \arg \max \left(\text{score}_{MNTD\text{-}PGD}, \ \text{score}_{SG\text{-}PGD}\right).
\end{align}
By coupling perceptual similarity with misclassification success, the Selection Module enforces a balance between transferability and visual fidelity.

\section{Experiments}

\paragraph{Test Set.}
We use the test set provided by the \textit{AADD Challenge: ACM Multimedia 2025} \cite{bongini2025wildnewinthewildimage} to validate our attack strategy. The dataset contains 693 high-quality (HQ) and 710 low-quality (LQ) deepfake images generated using the models in Table \ref{tab:generators}. A typical deepfake image is a human portrait with varied expressions and complex backgrounds. The number of images generated by GAN and Diffusion Models is numerically balanced. The LQ images are created by resizing followed by variable Quality Factor (QF) compression. This combination is designed to simulate social media compression.

\begin{figure}[ht]
    \centering
    \begin{tabular}{ccccc}
        \includegraphics[width=0.15\textwidth]{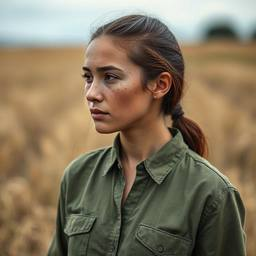} &
        \includegraphics[width=0.15\textwidth]{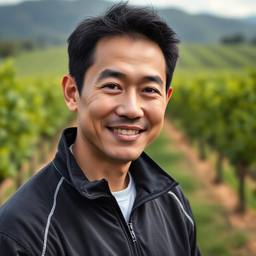} &
        \includegraphics[width=0.15\textwidth]{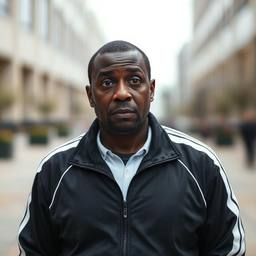} &
        \includegraphics[width=0.15\textwidth]{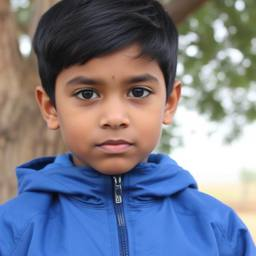} &
        \includegraphics[width=0.15\textwidth]{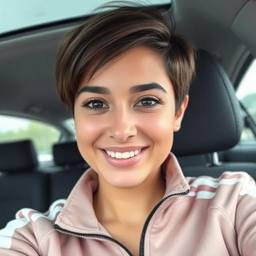} \\
        \includegraphics[width=0.15\textwidth]{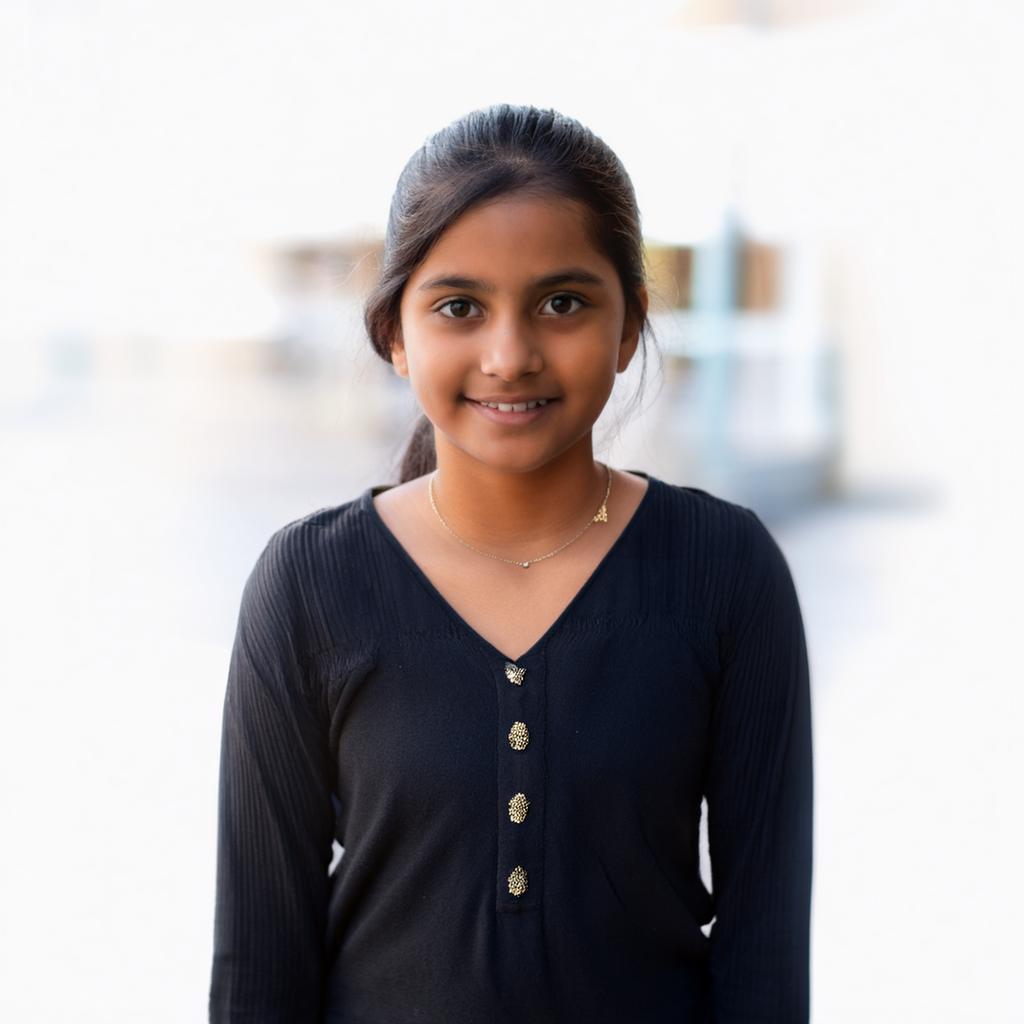} &
        \includegraphics[width=0.15\textwidth]{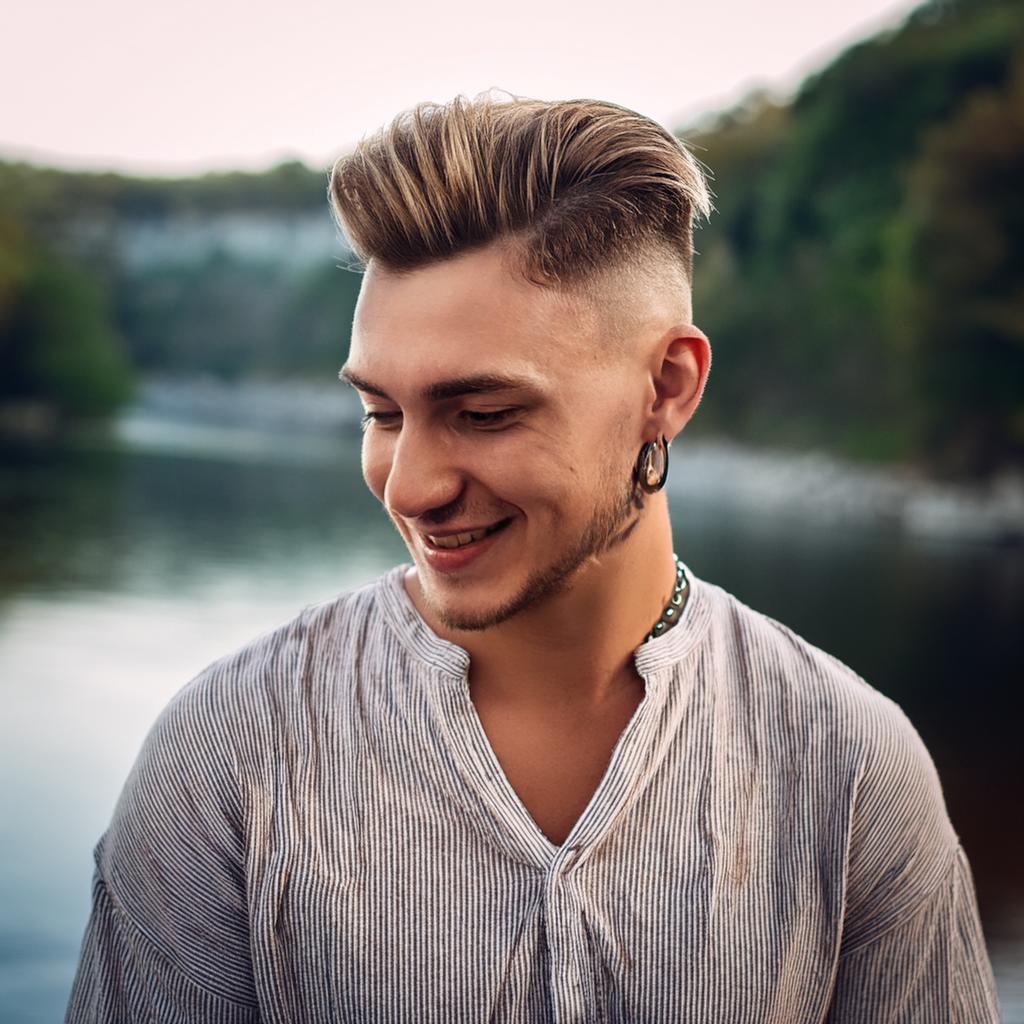} &
        \includegraphics[width=0.15\textwidth]{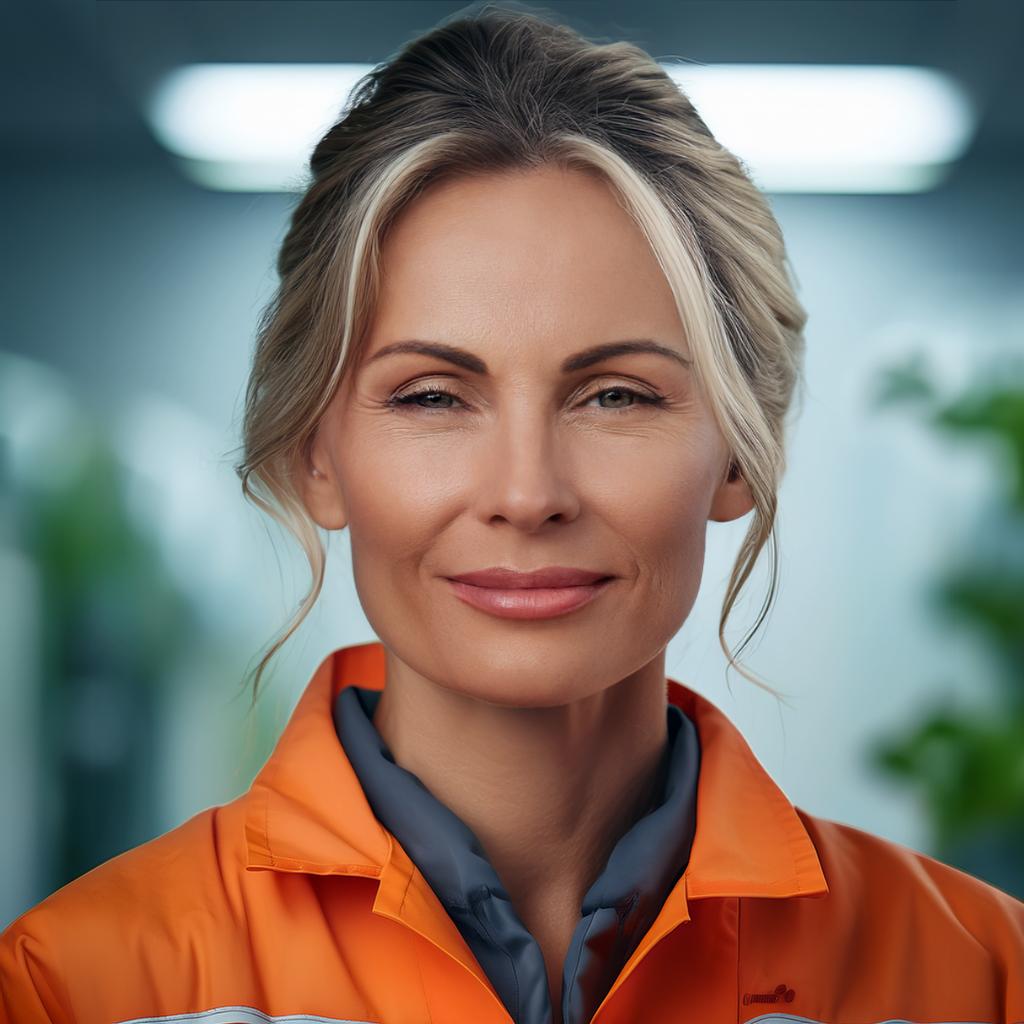} &
        \includegraphics[width=0.15\textwidth]{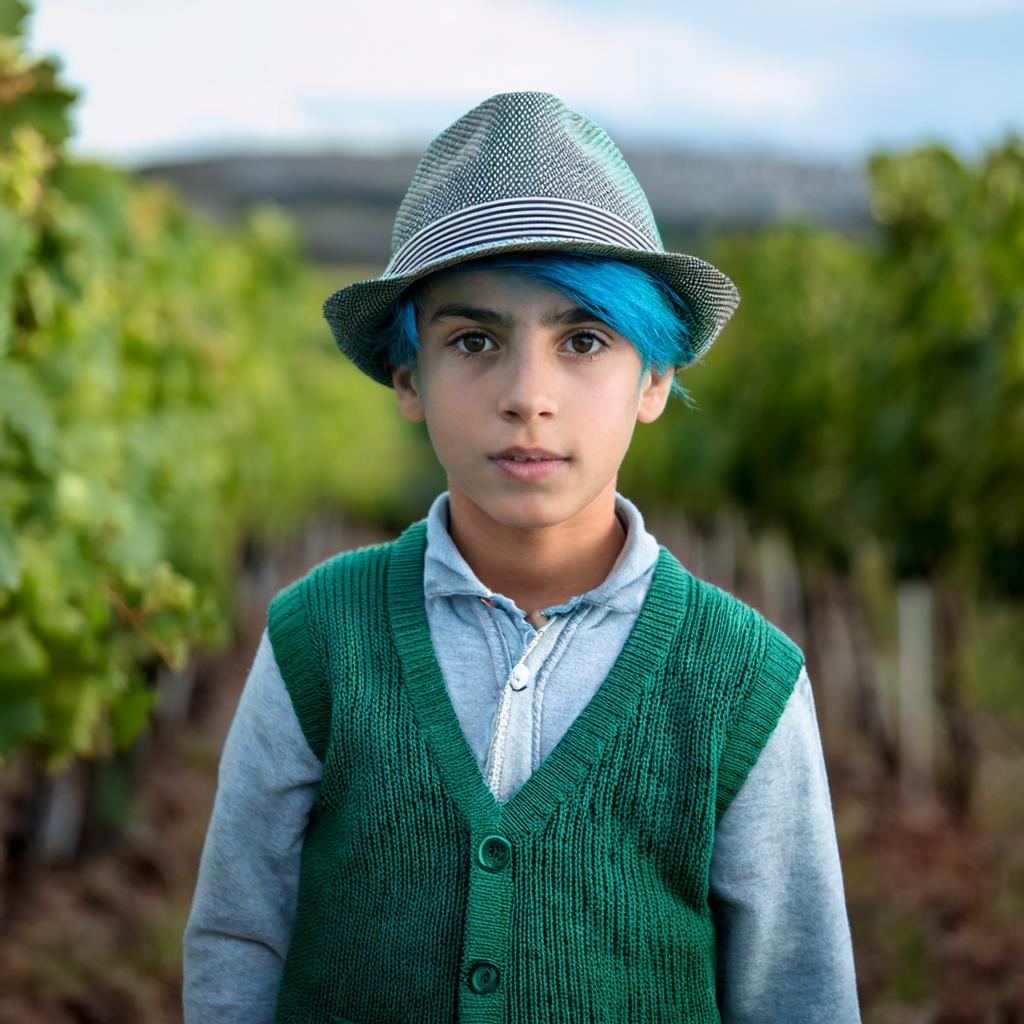} &
        \includegraphics[width=0.15\textwidth]{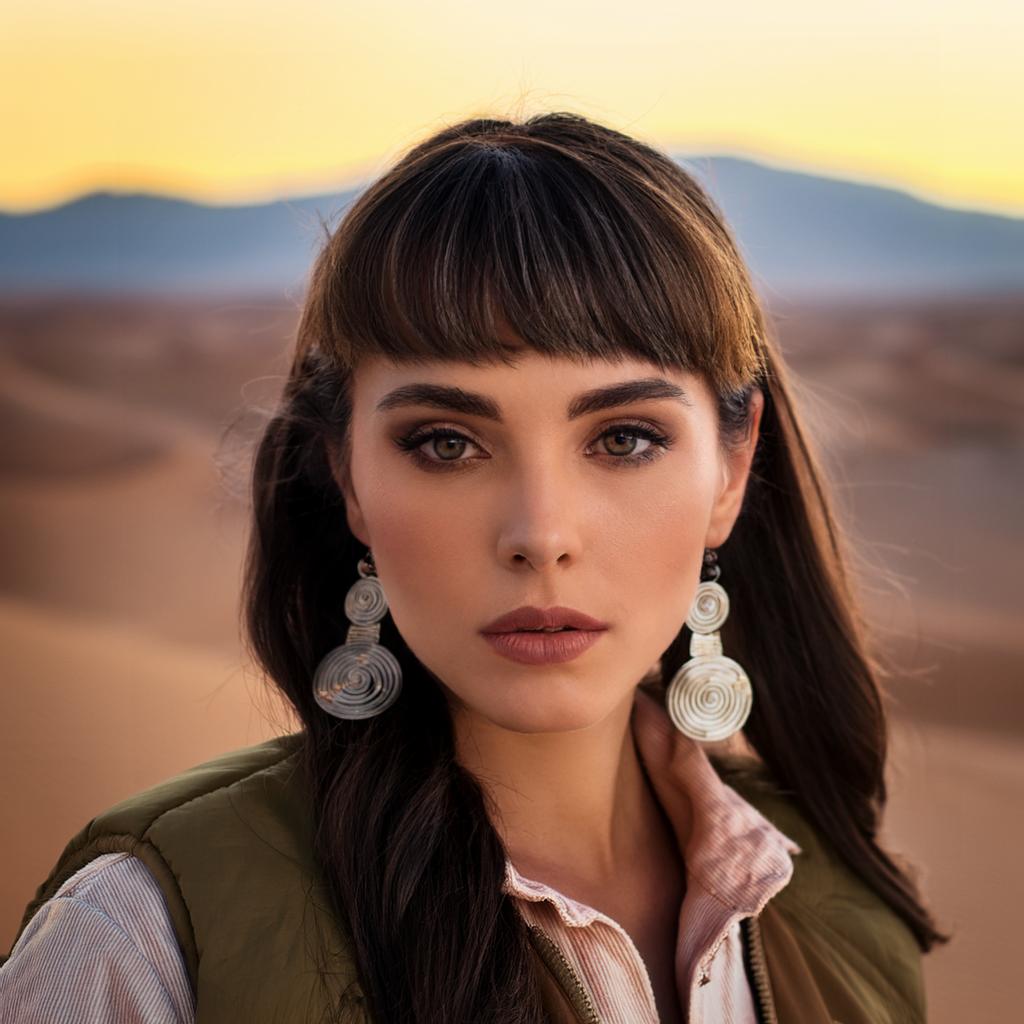}
    \end{tabular}
    \vspace{1em}
    \caption{\textbf{Top}: Low-quality deepfakes; \textbf{Bottom}: High-quality deepfakes.}
    \label{fig:my_image_grid}
\end{figure}

\begin{table}[ht]
\centering
\begin{tabular}{|l|p{10cm}|}
\hline
\textbf{Quality} & \textbf{Deepfake Generators} \\
\hline
HQ & Adobe Firefly, DeepAI, Flux 1.1 Pro, HotPotAI, Nvidia SanaPAG, Stable Diffusion 3.5, StyleGAN 2, StyleGAN 3, Tencent Hunyuan \\
\hline
LQ & DeepAI, Flux.1, Freepik, HotPotAI, Nvidia SanaPAG, Stable Diffusion Attend and Excite, StyleGAN, StyleGAN 3, Tencent Hunyuan \\
\hline
\end{tabular}
\vspace{1em}
\caption{Deepfake generators used to create the test set.}
\label{tab:generators}
\end{table}

\paragraph{Target and Surrogate Models.}
We used two target models, ResNet-50 \cite{He_2016_CVPR} and DenseNet-121 \cite{Huang_2017_CVPR}, each configured as a binary classifier for deepfake detection in a black-box setting. In each experiment, one model served as the target while the other acted as a surrogate. To further increase architectural diversity, improve transferability, and reduce overfitting, we incorporated two additional surrogate models, EfficientNet-B0 \cite{tan2019efficientnet} and Inception-v3 \cite{szegedy2016rethinking}, as robust deepfake detectors.

\paragraph{Baselines.}
We compare MS-GAGA against the Carlini \& Wagner $L_\infty$ attack \cite{CW2018} (white-box) and the Square Attack \cite{squareattack} (black-box), both implemented via the Adversarial Robustness Toolbox (ART) \cite{art2018}. While more advanced black-box attacks such as GRAPHITE \cite{graphite} exist, they are computationally intensive (require days of runtime compared to the hours needed for the other methods), thereby not offering a practical or fair comparison.

All attack strategies are evaluated using the following metrics:
\begin{itemize}[noitemsep, topsep=1pt]
    \item \textbf{Misclassification Rate:} Number of times an adversarial deepfake image successfully fools the target deepfake detector.
    \item \textbf{SSIM:} Computed using \texttt{skimage.metrics.structural\_similarity} with default window size=7 on each RGB channel.
\end{itemize}

\section{Results}

\begin{table}[ht]
\centering
\begin{tabular}{|c|c|c|c|}
\hline
\textbf{Original} & \textbf{MS-GAGA} & \textbf{Square Attack} & \textbf{Carlini-Wagner} \\
\toprule

\begin{minipage}{0.125\textwidth}
\centering
\includegraphics[width=\linewidth]{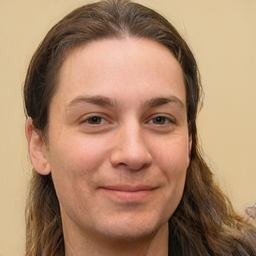}
\small SSIM: 1.0
\end{minipage} &

\begin{minipage}{0.125\textwidth}
\centering
\includegraphics[width=\linewidth]{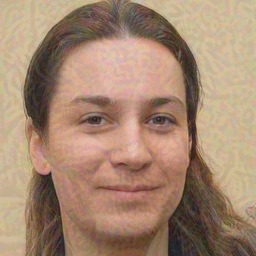}\\
\small SSIM: 0.83
\end{minipage} &

\begin{minipage}{0.125\textwidth}
\centering
\includegraphics[width=\linewidth]{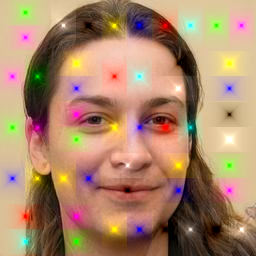}\\
\small SSIM: 0.78
\end{minipage} &

\begin{minipage}{0.125\textwidth}
\centering
\includegraphics[width=\linewidth]{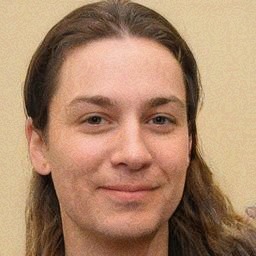}\\
\small SSIM: 0.86
\end{minipage} \\

\midrule

\begin{minipage}{0.125\textwidth}
\centering
\includegraphics[width=\linewidth]{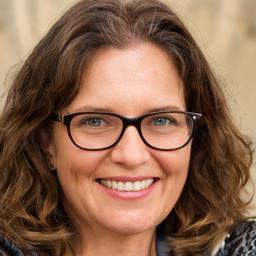}
\small SSIM: 1.0
\end{minipage} &

\begin{minipage}{0.125\textwidth}
\centering
\includegraphics[width=\linewidth]{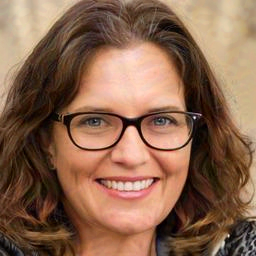}\\
\small SSIM: 0.95
\end{minipage} &

\begin{minipage}{0.125\textwidth}
\centering
\includegraphics[width=\linewidth]{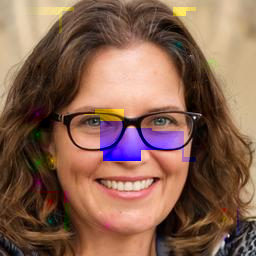}\\
\small SSIM: NA
\end{minipage} &

\begin{minipage}{0.125\textwidth}
\centering
\includegraphics[width=\linewidth]{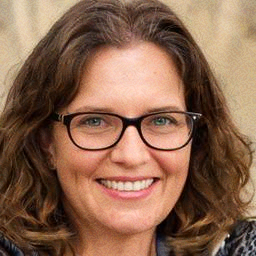}\\
\small SSIM: 0.93
\end{minipage} \\

\midrule

\begin{minipage}{0.125\textwidth}
\centering
\includegraphics[width=\linewidth]{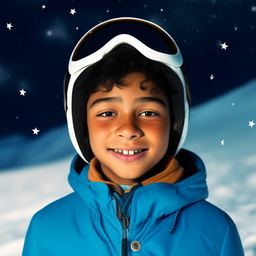}
\small SSIM: 1.0
\end{minipage} &

\begin{minipage}{0.125\textwidth}
\centering
\includegraphics[width=\linewidth]{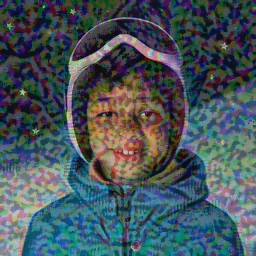}\\
\small SSIM: 0.27
\end{minipage} &

\begin{minipage}{0.125\textwidth}
\centering
\includegraphics[width=\linewidth]{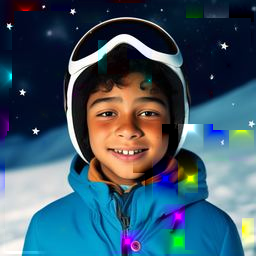}\\
\small SSIM: NA
\end{minipage} &

\begin{minipage}{0.125\textwidth}
\centering
\includegraphics[width=\linewidth]{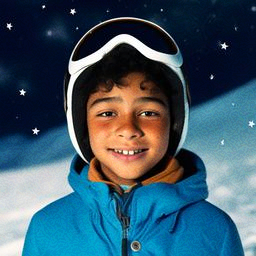}\\
\small SSIM: NA
\end{minipage} \\

\bottomrule
\end{tabular}
\vspace{1em}
\caption{Comparison of generated adversarial deepfake images. An SSIM of ``NA" indicates that the perturbed image failed to fool the classifier. In the first and third rows, MS-GAGA images have $0.8x$ contrast of the original.}
\label{fig:attack_comparison}
\vspace{-1em}
\end{table}

\begin{table}[t!]
\centering
\begin{tabular}{lcccc}
\toprule
\multirow{2}{*}{\textbf{Method}} & \multicolumn{2}{c}{\textbf{Misclassification Rate (\%)}} & \multirow{2}{*}{\textbf{Avg. SSIM}} & \multirow{2}{*}{\textbf{Score}} \\
\cmidrule(lr){2-3}
& \textit{ResNet-50} & \textit{DenseNet-121} & \\
\midrule
No Attack         & 0 & 0 & 100 & 0\\
Square Attack \cite{squareattack} & 56.3  & 56.6  & 73.4 & 1163 \\
Carlini-Wagner Attack \cite{CW2018} & 73.7 & 71.0  &  \textbf{91.3} & 1855\\
MS-GAGA (Ours) & \textbf{99.6} & \textbf{98.8} & 69.0 & \textbf{1921} \\
\bottomrule
\end{tabular}
\vspace{1em}
\caption{Performance comparison of various attacks}
\label{tab:results}
\vspace{-1.5em}
\end{table}

\paragraph{Quantitative Analysis.}
Table \ref{fig:attack_comparison} and \ref{tab:results} detail our experimental results. MS-GAGA achieves a 27\% higher misclassification rate than the other methods in all scenarios. However, it comes at the cost of lower SSIM. To weigh the trade-offs, we calculate the score on the entire test set using (\ref{eqn: score}), which shows that MS-GAGA still achieves better results overall.

The adversarial images generated by MS-GAGA are transferable, at least between the similar architectures of ResNet-50 and DenseNet-121. In contrast, the Carlini-Wagner attack tends to overfit the surrogate model, thereby lacking transferability. Additionally, the Square attack is innately less powerful than gradient-based methods.

\begin{figure}[th]
\centering
\begin{tabular}{c}
\includegraphics[width=.95\linewidth]{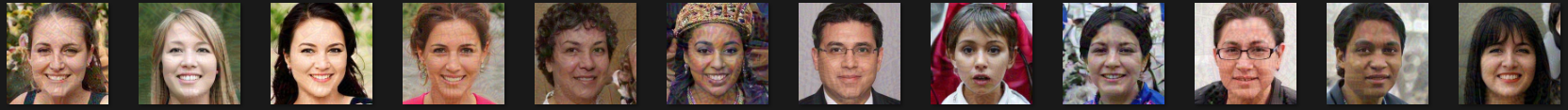}\\
\includegraphics[width=.95\linewidth]{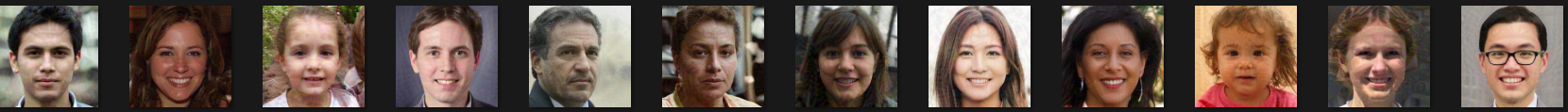}\\
\end{tabular}
\vspace{0.5em}
\caption{\textbf{Top}: MNTD-PGD ($SSIM \in [0.8-0.9]$); \textbf{Bottom}: SG-PGD ($SSIM \in [0.9-0.1)$)}
\label{fig:ablation}
\end{figure}

\paragraph{Ablation Study.}
To better understand the contribution of each stream in MS-GAGA, we conduct an ablation study where MNTD-PGD and SG-PGD are evaluated independently. Table~\ref{tab:ablation} reports their quantitative performance, while Figure~\ref{fig:ablation} provides visual examples of the resulting adversarial perturbations at different SSIM ranges.

\begin{table}[th]
\centering
\begin{tabular}{lcccc}
\toprule
\multirow{2}{*}{\textbf{Method}} & \multicolumn{2}{c}{\textbf{Misclassification Rate (\%)}} & \multirow{2}{*}{\textbf{Avg. SSIM}} & \multirow{2}{*}{\textbf{Score}} \\
\cmidrule(lr){2-3}
& \textit{ResNet-50} & \textit{DenseNet-121} & \\
\midrule
MNTD-PGD \ref{mntd-pgd}      & \textbf{98.2} & \textbf{97.6} & 67.5 & \textbf{1854}\\
SG-PGD   \ref{sg-pgd}        & 43.8 & 35.1 & \textbf{82.1} & 909 \\
\bottomrule
\end{tabular}
\vspace{1em}
\caption{MNTD-PGD provides transferability, SG-PGD aids higher structural similarity.}
\label{tab:ablation}
\end{table}
\vspace{-2em}

\vspace{-2em}
\section{Conclusion}

We introduce MS-GAGA, a two-stage adversarial framework designed to generate highly transferable attacks against black-box deepfake detectors. The results highlight significant vulnerabilities in current deepfake detection pipelines, demonstrating the need for more resilient defences. While this study successfully moves beyond single-surrogate and white-box limitations, it is confined to CNN-based models and static image datasets. Future work will extend this framework to additional deepfake models and modalities and will evaluate its effectiveness against varied detector architectures, such as transformer-based and frequency-domain models, to provide a more comprehensive understanding of its generalizability. This research provides a critical step toward safeguarding digital media integrity by offering a rigorous method to evaluate and strengthen detection systems.


\section*{Acknowledgment}
This research is supported by the Ministry of Education, Singapore, under its Academic Research Tier 1 (Grant number: GMS 956).

\bibliography{egbib}
\end{document}